\documentclass[lettersize,journal]{IEEEtran}
\usepackage{amsmath,amsfonts}
\usepackage{algorithmic}
\usepackage{algorithm}
\usepackage{array}
\usepackage[caption=false,font=normalsize,labelfont=sf,textfont=sf]{subfig}
\usepackage{textcomp}
\usepackage{stfloats}
\usepackage{url}
\usepackage{verbatim}
\usepackage{graphicx}
\usepackage{cite}
\usepackage{amssymb}
\usepackage{mathtools}
\usepackage{amsthm}
\usepackage{makecell}
\usepackage{booktabs} 

\newtheorem{theorem}{Theorem}[section]
\newtheorem{proposition}[theorem]{Proposition}

\usepackage{xcolor}

\begin{document}

\title{Restricted Generative Projection for One-Class Classification and Anomaly Detection}

\author{Feng Xiao, Ruoyu Sun, Jicong Fan~\IEEEmembership{Member,~IEEE,}
\thanks{The authors are with the School of Data Science, The Chinese University of Hong Kong, Shenzhen, and Shenzhen Research Institute of Big Data. E-mail: fanjicong@cuhk.edu.cn.}
\thanks{Manuscript received April 19, 2021; revised August 16, 2021.}}

\markboth{Journal of \LaTeX\ Class Files,~Vol.~14, No.~8, August~2021}%
{Shell \MakeLowercase{\textit{et al.}}: A Sample Article Using IEEEtran.cls for IEEE Journals}


\maketitle

\begin{abstract}
We present a simple framework for one-class classification and anomaly detection. The core idea is to learn a mapping to transform the unknown distribution of training (normal) data to a known target distribution. Crucially, the target distribution should be sufficiently simple, compact, and informative. The simplicity is to ensure that we can sample from the distribution easily, the compactness is to ensure that the decision boundary between normal data and abnormal data is clear and reliable, and the informativeness is to ensure that the transformed data preserve the important information of the original data. Therefore, we propose to use truncated Gaussian, uniform in hypersphere, uniform on hypersphere, or uniform between hyperspheres, as the target distribution.  We then minimize the distance between the transformed data distribution and the target distribution while keeping the reconstruction error for the original data small enough. Comparative studies on multiple benchmark datasets verify the effectiveness of our methods in comparison to baselines.
\end{abstract}


\section{Introduction}
\IEEEPARstart{A}{nomaly} detection (AD) under the setting of one-class classification aims to distinguish normal data and abnormal data using a model trained on only normal data \cite{chandola2009anomaly, pang2021deep, ruff2021unifying}. AD is useful in numerous real problems such as intrusion detection for video surveillance, fraud detection in finance, and fault detection for sensors. Many AD methods have been proposed in the past decades \cite{scholkopf1999support, scholkopf2001estimating,tax2004support, liu2008isolation, hu2018anomaly}. For instance, Sch{\"o}lkopf et al.\cite{scholkopf2001estimating} proposed the one-class support vector machine (OC-SVM) that finds, in a high-dimensional kernel feature space, a hyperplane yielding a large distance between the normal training data and the origin. Tax et al.\cite{tax2004support} presented the support vector data description (SVDD), which obtains a spherically shaped boundary (with minimum volume) around the normal training data to identify abnormal samples. Hu et al.\cite{hu2018anomaly}  propose a new kernel function to estimate samples’ local densities and propose a weighted
neighborhood density estimation to increase the robustness to changes in the neighborhood size.
There are also many deep learning based AD methods including unsupervised AD methods \cite{liu2021anomaly, ruff2018deep, golan2018deep, zong2018deep,wang2019multivariate,  qiu2021neural, huang2021hybrid} and semi-supervised AD methods \cite{hendrycks2018deep, ruff2019deep, liznerski2020explainable, ruff2020rethinking}.

Deep learning based AD methods may be organized into three categories. The first category is based on compression and reconstruction. These methods usually use an autoencoder \cite{hinton2006reducing,kingma2013auto} to learn a low-dimensional representation to reconstruct the high-dimensional data \cite{vincent2008extracting,wang2021auto}. The autoencoder learned from the normal training data is expected to have a much higher reconstruction error on unknown abnormal data than on normal data.
The second category is based on the combination of classical one-class classification \cite{tax2004support, golan2018deep} and deep learning \cite{ruff2018deep, ruff2019deep, ruff2020rethinking, perera2019learning, bhattacharya2021fast,shenkar2022anomaly,chen2022deep}. For instance, Ruff et al.\cite{ruff2018deep} proposed a method called deep one-class SVDD. The main idea is to use deep learning to construct a minimum-radius hypersphere to include all the training data, while the unknown abnormal data are expected to fall outside. 
The last category is based on generative learning or adversarial learning
\cite{malhotra2016lstm,deecke2018image,pidhorskyi2018generative, pmlr-v97-nguyen19b,perera2019ocgan, goyal2020drocc, raghuram2021general,yan2021learning, zheng2021generative}. 
For example, Perera et al. \cite{perera2019ocgan} proposed to use the generative adversarial network (GAN) \cite{goodfellow2014generative} with constrained latent representation to detect anomalies for image data. Goyal et al.\cite{goyal2020drocc} presented a method called deep robust one-class classification (DROCC) and the method aims to find a low-dimensional manifold to accommodate the normal data via an adversarial optimization approach. 

Although deep learning based AD methods have shown promising performance on various datasets, they still have limitations. For instance, the one-class classification methods such as Deep SVDD \cite{ruff2018deep} only ensure that a hypersphere could include the normal data but cannot guarantee that the normal data are distributed evenly in the hypersphere, which may lead to large empty regions in the hypersphere and hence yield incorrect decision boundary (see Fig.\ref{fig1}). Moreover, the popular hypersphere assumption may not be the best one for providing a compact decision boundary (see Fig.\ref{fig2} and Tab.\ref{tab:density}). The adversarial learning methods such as \cite{pmlr-v97-nguyen19b,perera2019ocgan, goyal2020drocc, du2021gan} may suffer from instability in optimization. 

In this work, we present a restricted generative projection (RGP) framework for one-class classification and anomaly detection. The main idea is to train a deep neural network to convert the distribution of normal training data to a target distribution that is simple, compact, and informative, which will provide a reliable decision boundary to identify abnormal data from normal data. There are many choices for the target distribution, such as truncated Gaussian and uniform on hypersphere. Our contributions are summarized as follows.
\begin{itemize}
    \item We present a novel framework called RGP for one-class classification and anomaly detection. It aims to transform the data distribution to some target distributions that are easy to be violated by unknown abnormal data.
    \item We provide four simple, compact, and informative target distributions, analyze their properties theoretically, and show how to sample from them efficiently. 
    \item We propose two extensions for our original RGP method.
\end{itemize}
We conduct extensive experiments (on eight benchmark datasets) to compare the performance of different target distributions and compare our method with state-of-the-art baselines. The results verify the effectiveness of our methods.
The rest of this paper is organized as follows. Section \ref{sec:relatedwork} introduces the related work.
Section \ref{section3} details our proposed methods.
Section \ref{sec:extensions} presents two extensions of the proposed method.
Section \ref{experiments} shows the experiments.
Section \ref{conclusion} draws conclusions for this paper.

\section{Related Work}\label{sec:relatedwork}
Before elaborating our method, we in this section briefly review deep one-class classification, autoencoder-based AD methods, and maximum mean discrepancy (MMD)\cite{gretton2012kernel}. 
We also discuss the connection and difference between our method and these related works. 

\subsection{Deep One-Class Classification}
The Deep SVDD proposed by \cite{ruff2018deep} uses a neural network to learn a minimum-radius hypersphere to enclose the normal training data, i.e.,
\begin{equation}\label{eq_deepsvdd}
    \mathop{\text{minimize}}_{\mathcal{W}} \frac{1}{n}\sum^n_{i=1} \Vert \phi(\mathbf{x}_i; \mathcal{W}) - \mathbf{c} \Vert^2 + \frac{\lambda}{2}\sum^L_{l=1}\Vert \mathbf{W}_l \Vert^2_F
\end{equation}
where $\mathbf{c} \in \mathbb{R}^d$ is a predefined centroid and $\mathcal{W}=\{\mathbf{W}_1,\ldots,\mathbf{W}_L\}$ denotes the parameters of the $L$-layer neural network $\phi$,  and $\lambda$ is a regularization hyperparameter. In \eqref{eq_deepsvdd}, to avoid model collapse, bias terms should not be used and activation functions should be bounded \cite{ruff2018deep}. There are also a few variants of Deep SVDD proposed for semi-supervised one-class classification and anomaly detection \cite{ruff2019deep, ruff2020rethinking}.

 \begin{figure}[h]
\centering
    \includegraphics[width=0.90\linewidth]{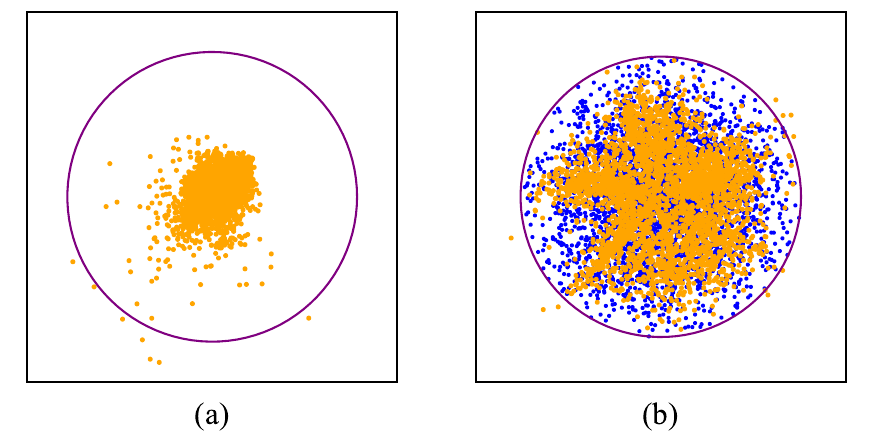}
\caption{Visualization of transforming the training data of Thyroid dataset (detailed in Section \ref{sec_data_baseline}) to a 2-dimensional space. Plots (a) and (b) correspond to Deep SVDD and our method respectively. The orange points denote the transformed normal training data and the purple circles denote the decision boundaries. The blue points in plot (b) denote the samples drawn from a truncated Gaussian. In plot (a), the hypersphere is not a good boundary to describe the distribution of the normal data, while in plot (b), the hypersphere is good enough to describe the distribution of the normal data. Particularly, in plot (a), if we reduce the radius of the hypersphere, there will be many normal data points falling outsides the decision boundary, which contradicts the assumption that all or last least most of the training data are normal.}\label{fig1}
\end{figure}

Both our method and Deep SVDD as well as its variants aim to project the normal training data into some space such that a decision boundary between normal data and unknown abnormal data can be found easily. However, the sum-of-square minimization in Deep SVDD and its variants only ensures that the projected data are sufficiently close to the centroid $\mathbf{c}$ in the sense of Euclidean distance and does guarantee that the data are sufficiently or evenly distributed in the hypersphere centered at $\mathbf{c}$. Thus, in the hypersphere, there could be holes or big empty regions without containing any normal data and hence it is not suitable to assume that the whole space enclosed by the hypersphere is completely a normal space. In other words, the optimal decision boundary between normal data and abnormal data is actually very different from the hypersphere. An intuitive example is shown in Fig.\ref{fig1}. We see that there is a large empty space in the hypersphere learned by Deep SVDD. In contrast, the transformed data of our method are sufficiently distributed. 

\subsection{Autoencoder-based AD Methods}

Our method is similar to but quite different from the variational autoencoder (VAE) \cite{kingma2013auto}. Although our model is an autoencoder, the main goal is not to represent or generate data; instead, our model aims to convert distribution to find a reliable decision boundary for anomaly detection.  More importantly, the latent distribution in VAE is often Gaussian and not bounded while the latent distribution in our model is more general and bounded, which is essential for anomaly detection. In addition, the optimizations of VAE and our method are also different: VAE involves KL-divergence while our method involves maximum mean discrepancy \cite{gretton2012kernel}.

It is worth noting that similar to our method, Perera et al.\cite{perera2019ocgan} also considered bounded latent distribution in autoencoder for anomaly detection. They proposed to train a denoising autoencoder with a hyper-cube supported latent space, via adversarial training. The latent distribution and optimization are different from ours. In addition, the latent distributions of our method, such as uniform on hypersphere, are more compact than the multi-dimensional uniform latent distribution of their method. 

Compared with the autoencoder based anomaly detection method NAE \cite{yoon2021autoencoding} that uses reconstruction error to normalize autoencoder, our method pays more attention to learning a mapping that can transform the unknown data distribution into a simple and compact target distribution. The ideas are orthogonal.

\subsection{Maximum Mean Discrepancy}
In statistics, maximum mean discrepancy (MMD)\cite{gretton2012kernel} is often used for Two-Sample test and its principle is to find a function that assumes different expectations on two different distributions: 
\begin{equation}
    \label{eq3}
          \text{MMD}[\mathcal{F}, p,q] =\underset{\Vert f \Vert_{\mathcal{H}}\leq1}{\sup}\left(\mathbb{E}_p[f(\mathbf{x})]-\mathbb{E}_q[f(\mathbf{y})]\right),
\end{equation}
where $p, q$ are probability distributions, $\mathcal{F}$ is a class of functions $ f:\mathbb{X} \rightarrow \mathbb{R}$ and $\mathcal{H}$ denotes a reproducing kernel Hilbert space.
Using the kernel trick, MMD can be represented as a simple loss function to measure the discrepancy between two distributions by finite samples, which is easy to apply to deep learning and can be efficiently trained by gradient descent. Based on the aforementioned advantages of MMD, Li et al.\cite{li2015generative} proposed generative moment matching networks (GMMNs), which leads to a simpler optimization objective compared to the min-max optimization of GAN \cite{goodfellow2014generative}.

Although both our method and GMMNs \cite{li2015generative} minimize the MMD between data distribution and prior distribution, our goal is not generating new data but detecting anomalies. In addition, we consider a few bounded target distributions and analyze their sampling properties. More importantly, our method has very competitive performance when compared with SOTA methods of anomaly detection and one-class classification.

\section{Restricted Generative Projection}
\label{section3}

In this section, we introduce our RGP framework, bounded target distributions, and the computation of anomaly scores.
\subsection{Restricted Distribution Projection}

Suppose we have a set of $m$-dimensional training data $\mathbf{X}=\{\mathbf{x}_1,\mathbf{x}_2,\ldots,\mathbf{x}_n \}$
drawn from an unknown bounded distribution $\mathcal{D}_{\mathbf{x}}$ and any samples drawn from $\mathcal{D}_{\mathbf{x}}$ are normal data. We want to train a model $\mathcal{M}$ on $\mathbf{X}$ to determine whether a test data $\mathbf{x}_{\text{new}}$ is drawn from $\mathcal{D}_{\mathbf{x}}$ or not. One may consider estimating the density function (denoted by ${p}_{\mathbf{x}}$) of $\mathcal{D}_{\mathbf{x}}$ using some techniques such as kernel density estimation \cite{rosenblatt1956remarks}. Suppose the estimation $\hat{p}_{\mathbf{x}}$ is good enough, then one can determine whether $\mathbf{x}_{\text{new}}$ is normal or not according to the value of $\hat{p}_{\mathbf{x}}(\mathbf{x}_{\text{new}})$: if $\hat{p}_{\mathbf{x}}(\mathbf{x}_{\text{new}})$ is zero or close to zero, $\mathbf{x}_{\text{new}}$ is an abnormal data point; otherwise, $\mathbf{x}_{\text{new}}$ is a normal data point \footnote{Here we assume that the distributions of normal data and abnormal data do not overlap. Otherwise, it is difficult to determine whether a single point is normal or not.}. However, the dimensionality of the data is often high and hence it is very difficult to obtain a good estimation $\hat{p}_{\mathbf{x}}$.

We propose to learn a mapping $\mathcal{T}:\mathbb{R}^m\rightarrow\mathbb{R}^d$ to transform the unknown bounded distribution $\mathcal{D}_{\mathbf{x}}$ to a known distribution $\mathcal{D}_{\mathbf{z}}$ while there still exists a mapping $\mathcal{T}':\mathbb{R}^d\rightarrow\mathbb{R}^m$ that can recover $\mathcal{D}_{\mathbf{x}}$ from $\mathcal{D}_{\mathbf{z}}$ approximately.
Let $p_{\mathbf{z}}$ be the density function of $\mathcal{D}_{\mathbf{z}}$. Then we can determine whether $\mathbf{x}_{\text{new}}$ is normal or not according to the value of $p_{\mathbf{z}}(\mathcal{T}(\mathbf{x}_{\text{new}}))$. To be more precise, we want to solve the following problem
\begin{equation}    
    \label{eq0}
          \underset{\mathcal{T},~\mathcal{T'}}{\text{minimize}} ~\mathcal{M}\left(\mathcal{T}(\mathcal{D}_{\mathbf{x}}), \mathcal{D}_{\mathbf{z}}\right)+\lambda \mathcal{M}\left(\mathcal{T}'(\mathcal{T}(\mathcal{D}_{\mathbf{x}})),\mathcal{D}_{\mathbf{x}}\right),
\end{equation}
where $\mathcal{M}(\cdot, \cdot)$ denotes some distance metric between two distributions and $\lambda$ is a trade-off parameter for the two terms. Note that if $\lambda=0$, $\mathcal{T}$ may convert any distribution to $\mathcal{D}_{\mathbf{z}}$ and lose the ability of distinguishing normal data and abnormal data.
Based on the universal approximation theorems \cite{pinkus1999approximation,lu2017expressive} and substantial success of neural networks, we use deep neural networks (DNN) to model $\mathcal{T}$ and $\mathcal{T}'$ respectively.  Let $f_{\theta}$ and $g_{\phi}$ be two DNNs with parameters $\theta$ and $\phi$ respectively. We solve 
\begin{equation}    
    \label{eq1}
          \underset{\theta,~\phi}{\text{minimize}} ~\mathcal{M}\left(\mathcal{D}_{f_\theta(\mathbf{x})}, \mathcal{D}_{\mathbf{z}}\right)+\lambda \mathcal{M}\left(\mathcal{D}_{g_\phi(f_\theta(\mathbf{x}))}, \mathcal{D}_{\mathbf{x}}\right),
\end{equation}
where $f_{\theta}$ and $g_\phi$ serve as encoder and decoder respectively.
However, problem (\ref{eq1}) is intractable because $\mathcal{D}_{\mathbf{x}}$ is unknown and $\mathcal{D}_{f_\theta(\mathbf{x})}$, $\mathcal{D}_{g_\phi(f_\theta(\mathbf{x}))}$ cannot be computed analytically. Note that the samples of $\mathcal{D}_{\mathbf{x}}$ and $\mathcal{D}_{g_\phi(f_\theta(\mathbf{x}))}$ are given and paired. Then the second term in the objective of (\ref{eq1}) can be replaced by sample reconstruction error such as $\tfrac{1}{n}\sum_{i=1}^n\|\mathbf{x}_i-g_\phi(f_\theta(\mathbf{x}_i))\|^2$. On the other hand, we can also sample from $\mathcal{D}_{f_\theta(\mathbf{x})}$ and $\mathcal{D}_{\mathbf{z}}$ easily but their samples are not paired. Hence, the metric $\mathcal{M}$ in the first term of the objective of (\ref{eq1}) should be able to measure the distance between two distributions using their finite samples. To this end, we propose to use the kernel maximum mean discrepancy (MMD)\cite{gretton2012kernel} to measure the distance between $\mathcal{D}_{f_\theta(\mathbf{x})}$ and $\mathcal{D}_{\mathbf{z}}$.
Its empirical estimate is
\begin{equation}
    \label{eq4}
        \begin{aligned}
          &\text{MMD}^2[\mathcal{F}, X,Y] = \frac{1}{m(m-1)}\sum_{i=1}^m\underset{j\neq i}{\sum^m}k(\mathbf{x}_i, \mathbf{x}_j) \\
          &+ \frac{1}{n(n-1)}\sum_{i=1}^n\underset{j\neq i}{\sum^n}k(\mathbf{y}_i, \mathbf{y}_j)
          - \frac{2}{mn}\sum_{i=1}^m\underset{j=1}{\sum^n}k(\mathbf{x}_i, \mathbf{y}_j),
          \end{aligned}
\end{equation}
where $X = \{\mathbf{x}_1, \dots, \mathbf{x}_m\}$ and $Y = \{\mathbf{y}_1, \dots, \mathbf{y}_n\}$ are samples consisting of i.i.d observations drawn from $p$ and $q$, respectively. $k(\cdot, \cdot)$ denotes a kernel function, e.g., $k(\mathbf{x}, \mathbf{y})=\exp(-\gamma\|\mathbf{x}-\mathbf{y}\|^2)$, a Gaussian kernel.

Based on the above analysis, we obtain an approximation for (\ref{eq1}) as
\begin{equation}
    \label{eq5}
    \mathop{\text{minimize}}_{\theta,~\phi}~ \text{MMD}^2(\mathbf{Z}_{\theta},\mathbf{Z}_T)+ \frac{\lambda}{n}\sum_{i=1}^n\|\mathbf{x}_i-g_\phi(f_\theta(\mathbf{x}_i))\|^2, 
\end{equation}
where $\mathbf{Z}_{\theta}=\{f_{\theta}(\mathbf{x}_1),f_{\theta}(\mathbf{x}_2),\ldots,f_{\theta}(\mathbf{x}_n) \}$ and  $\mathbf{Z}_T=\{\mathbf{z}_i:\mathbf{z}_i\sim\mathcal{D}_{\mathbf{z}},~i=1,\ldots,n\}$.
The first term of the objective function in (\ref{eq5}) makes $f_\theta$ learn the mapping $\mathcal{T}$ from data distribution $\mathcal{D}_{\mathbf{x}}$ to target distribution $\mathcal{D}_{\mathbf{z}}$ and the second term ensures that $f_\theta$ can preserve the main information of observations provided that $\lambda$ is sufficiently large. 

\subsection{Bounded Target Distributions}
\begin{figure}[h]
\centering
    \includegraphics[width=0.95\linewidth]{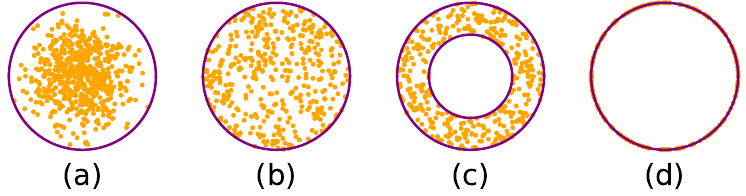}
\caption{Samples (in orange) from 2-D target (bounded) distributions. Plots (a), (b), (c), and (d) correspond to {GiHS}, {UiHS}, {UbHS}, and {UoHS} respectively. }\label{fig2}
\end{figure}

Now we introduce four examples of simple and compact $\mathcal{D}_{\mathbf{z}}$ for (\ref{eq5}). The four distributions are Gaussian in Hypersphere ({GiHS}),  Uniform in Hypersphere ({UiHS}), Uniform between Hyperspheres ({UbHS}), and
Uniform on Hypersphere ({UoHS}). Their 2-dimensional examples are visualized in Fig.\ref{fig2}.

\textbf{GiHS} (Fig.\ref{fig2}.a) is actually a truncated Gaussian. Suppose we want to draw $n$ samples from GiHS. A simple approach is drawing $(1+\rho)n$ samples from a standard $d$-dimensional Gaussian and discarding the $\rho n$ samples with larger $\ell_2$ norms. The maximum $\ell_2$ norm of the remaining $n$ points is the radius of the hypersphere. One may also use the inverse transform method of \cite{marsaglia1963generating}. We have the following results.
\begin{proposition}\label{prop_1}
Suppose $\mathbf{z}_1,\mathbf{z}_2,\ldots,\mathbf{z}_n$ are sampled from $\mathcal{N}(\mathbf{0},\mathbf{I}_d)$ independently. Then for any $r>\sqrt{d}$, we have
\begin{equation}\label{eq_prop1_a}
    \operatorname{Pr}\left(\|\mathbf{z}_j\| \geq r\right) \leq \exp \left(-0.5\alpha\right),\quad j\in[n],
\end{equation}
and 
\begin{equation}\label{eq_prop_1_b}
    \operatorname{Pr}\left(\max_{1\leq j\leq n}\Vert \mathbf{z}_j\Vert\leq r\right)\geq 1-n\exp \left(-0.5\alpha\right),
\end{equation}
where $\alpha=\sqrt{d+2r^2}-\sqrt{d}$.
\end{proposition}
Inequality \eqref{eq_prop_1_b} means a hypersphere of radius $r$ can include all the $n$ samples with a high probability if $r$ is sufficiently large.  On the other hand, according to \eqref{eq_prop1_a}, if we expect to get $n$ samples in a hypersphere of radius $r$, we need to sample about $n/(1-\exp(-0.5\alpha))$ points from $\mathcal{N}(\mathbf{0},\mathbf{I}_d)$. If $d$ is larger, we need to sample more points.

\textbf{UiHS} (Fig.\ref{fig2}.b) is a hyperball in which all the samples are distributed uniformly. To sample from UiHS, we first need to sample from $\mathcal{U}(-r,r)^{d}$. Then we discard all the data points outsides the radius-$r$ hyperball centered at the origin. 
The following proposition (the proof is in Appendix) shows some probability result of sampling from a $d$-dimensional uniform distribution.
\begin{proposition}\label{prop_2}
Suppose $\mathbf{z}_1,\mathbf{z}_2,\ldots,\mathbf{z}_n$ are sampled from $\mathcal{U}(-r,r)^{d}$ independently. Then for any $t>0$, we have
\begin{equation}\label{eq_prop2_a}
\operatorname{Pr}\left(\|\mathbf{z}_j\| \geq{rt}\right) \leq \frac{d}{3t^2},\quad j\in[n],
\end{equation}
and
\begin{equation}\label{eq_prop2_b}
       \operatorname{Pr}\left(\max_{1\leq j\leq n}\Vert \mathbf{z}_j\Vert\leq rt\right)\geq 1-\frac{nd}{3t^2}.
\end{equation}
\end{proposition}
Inequality \eqref{eq_prop2_b} means a hypersphere of radius $rt$ can include all the $n$ samples with probability at least $1-nd/(3t^2)$. On the other hand, inequality \eqref{eq_prop2_b} indicates that if we draw $n/(1-d/(3t^2))$ samples from $\mathcal{U}(-r,r)^{d}$, the expected number of samples falling into a hypersphere of radius $rt$ is at least $n$.
Actually, sampling from UiHS is closely related to the Curse of Dimensionality and we need to sample a large number of points from $\mathcal{U}(-r,r)^{d}$ if $d$ is large because only a small volume of the hypercube is inside the hyperball. To be more precisely, letting $V_{\mathrm {hypercube}}$ be the volume of a hypercube with length $2r$ and $V_{\mathrm {hyperball}}$ be the volume of a hyperball with radius $r$, we have
\begin{equation}
    \frac {V_{\mathrm {hyperball} }}{V_{\mathrm {hypercube} }}={\frac {\pi ^{d/2}}{d2^{d-1}\Gamma (d/2)}}\triangleq \eta,
\end{equation}
where $\Gamma$ is the gamma function. Therefore, we need to draw $n/\eta$ samples from $\mathcal{U}(-r,r)^{d}$ to ensure that the expected number of samples included in the hyperball is $n$, where $\eta$ is small if $d$ is large. 

\textbf{UbHS} (Fig.\ref{fig2}.c) can be obtained via UiHS. We first sample from UiHS and then remove all samples included by a smaller hypersphere. Since the volume ratio of two hyperballs with radius $r$ and $r'$is $(\frac{r}{r'})^d$, where $r'<r$, we need to draw $n/(1-(r'/r)^d)$ samples from UiHS to ensure that the expected number of samples between the two hyperspheres is $n$. Compared with GiHS and UiHS, UbHS is more compact and hence provides larger abnormal space for abnormal data to fall in.

\textbf{UoHS} (Fig.\ref{fig2}.d) can be easily obtained via sampling from $\mathcal{N}(\mathbf{0},\mathbf{I}_d)$. Specifically, for every $\mathbf{z}_i$ drawn from $\mathcal{N}(\mathbf{0},\mathbf{I}_d)$, we normalize it as $\mathbf{z}_i\leftarrow {r\mathbf{z}_i}/{\Vert \mathbf{z}_i\Vert}$, where $r$ is the predefined radius of the hypersphere. UoHS is a special case of UbHS when $r'=r$.

To quantify the compactness of the four target distributions, we define density $\rho$ as the number of data points in unit volume, i.e., $\rho=n/V$. Consequently, the densities of the four target distributions are reported in Table \ref{tab:density}.
UoHS is more compact than UbHS as well as GiHS and UiHS, it should have better performance in anomaly detection. Indeed, our numerical results show that UoHS outperforms others in most cases.

\begin{table}[h]
    \centering
        \caption{Densities of the four target distributions.}
    \label{tab:density}
    \begin{tabular}{c|c|c|c|c} \hline
        &GiHS & UiHS & UbHS & UoHS \\ \hline
        $\rho$& $\dfrac{n\Gamma(d/2+1)}{\pi^{d/2}r^d}$ & $\dfrac{n\Gamma(d/2+1)}{\pi^{d/2}r^d}$ & $\dfrac{n\Gamma(d/2+1)}{\pi^{d/2}(r^d-r'^d)}$ & $\infty$\\ \hline
    \end{tabular}
\end{table}

\subsection{Anomaly Scores}

In the test stage, we only use the trained $f_{\theta}^*$ to calculate anomaly scores. For a given test sample
$\mathbf{x}_{\text{new}}$, we define anomaly score $s$ for each target distribution by
\begin{equation}
    \label{eq12}
    s(\mathbf{x}_\text{new})= \left\{
    \begin{array}{l}
      |\Vert f_\theta^*(\mathbf{x}_\text{new}) \Vert - r |,\quad \text{for {UoHS}}\\
      \Vert f_\theta^*(\mathbf{x}_\text{new}) \Vert,\quad \text{for GiHB or UiHS}\\
        \left(\Vert f_\theta^*(\mathbf{x}_\text{new}) \Vert - r\right)\cdot (\Vert f_\theta^*(\mathbf{x}_\text{new}) \Vert - r'),\\ \qquad\qquad\qquad\qquad\qquad\qquad \text{for UbHS} 
    \end{array}\right.
\end{equation}

There are clear decision boundaries according to (\ref{eq12}) and they can be regarded as `hard boundaries' between normal samples and abnormal samples. However, these `hard boundaries' only work in ideal cases where the projected data exactly match the target distributions. In real cases, due to the noise of data or the non-optimality of optimization, the projected data do not exactly match the target distributions.  Therefore, we further propose a `soft boundary' for calculating anomaly scores. Specifically, for a given test sample $\mathbf{x}_{\text{new}}$, we define anomaly score $s$ for all four target distributions as
\begin{equation}
    \label{eq13}
    s(\mathbf{x}_\text{new})= \frac{1}{k}\sum_{i \in N_k} \Vert f_\theta^*(\mathbf{x}_\text{new}) - f_\theta^*(\mathbf{x}_i) \Vert
\end{equation}
where $\mathbf{x}_i$ denotes a single sample with index $i$ in the training data and $N_k$ denotes the index set of the $k$ nearest training (projected) samples to $f_\theta^*(\mathbf{x}_\text{new})$.

Empirically, in the experiments, we found that (\ref{eq13}) has better performance than (\ref{eq12}) in most cases. Table \ref{table1}, \ref{table2}, \ref{table3} only report the results from (\ref{eq13}). The comparison results between (\ref{eq12}) and (\ref{eq13}) are provided in Section \ref{section4.6}. 

We call our method Restricted Generative Projection (RGP), which has four variants, denoted by \textbf{RGP-GiHS}, \textbf{RGP-UiHS}, \textbf{RGP-UbHS}, and \textbf{RGP-UoHS} respectively, though any bounded target distribution applies.

\section{Extensions of RGP}\label{sec:extensions}

In this section, based on the general objective in \eqref{eq1}, we provide two variants of RGP.

\subsection{Double-MMD based RGP} 
In the objective function of RGP defined by \eqref{eq5}, the second term is the reconstruction error for $\mathbf{X}$, which is only a special example of approximation for the second term in the objective function of \eqref{eq1}, i.e., $\mathcal{M}\left(\mathcal{D}_{g_\phi(f_\theta(\mathbf{x}))}, \mathcal{D}_{\mathbf{x}}\right)$. Alternatively, we can use MMD to approximate $\mathcal{M}\left(\mathcal{D}_{g_\phi(f_\theta(\mathbf{x}))}, \mathcal{D}_{\mathbf{x}}\right)$, which yields the following \textbf{Double-MMD RGP}:
\begin{equation}
\label{eq15}
    \mathop{\text{minimize}}_{\theta,~\phi}~ \text{MMD}^2(\mathbf{Z}_{\theta},\mathbf{Z}_T)+ \lambda\text{MMD}^2(g_\phi(\mathbf{Z}_\theta),\mathbf{X}). 
\end{equation}
Compared to the sum of squares reconstruction error used in \eqref{eq5}, $\text{MMD}^2(g_\phi(\mathbf{Z}_\theta),\mathbf{X})$ is a weaker approximation for $\mathcal{M}\left(\mathcal{D}_{g_\phi(f_\theta(\mathbf{x}))}, \mathcal{D}_{\mathbf{x}}\right)$, 
because it does not exploit the fact that the samples in $\mathbf{Z}_{\theta}$ and $\mathbf{X}$ are paired. Thus, the projection of Double-MMD RGP cannot preserve sufficient information of $\mathbf{X}$, 
which will reduce the detection accuracy. Indeed, as shown by the experimental results in Section 
\ref{exp4.5}, our original RGP outperforms Double-MMD RGP.

\subsection{Sinkhorn Distance based RGP} 
Besides MMD, the optimal transport theory can also be used to construct a notion of distance between pairs of probability distributions. In particular, the Wasserstein distance \cite{kantorovich1960mathematical}, also known as “Earth Mover’s Distance”, has appealing theoretical properties and a very intuitive formulation
\begin{equation}
        \mathcal{W} = \langle \gamma^*, \mathbf{C} \rangle_F \\
\end{equation}
where $\mathbf{C}$ denotes a metric cost matrix and $\gamma*$\ is the optimal transport plan.
Finding the optimal transport plan $\gamma^*$ might appear to be a really hard problem. Especially, the computation cost of Wasserstein distance can quickly become prohibitive when the data dimension increases. In order to speed up the calculation of Wasserstein distance, Cuturi \cite{cuturi2013sinkhorn} proposed Sinkhorn distance that regularizes the optimal transport problem with an entropic penalty and uses Sinkhorn's algorithm \cite{sinkhorn1967concerning} to approximately calculate Wasserstein distance.

Now, if replacing the first term in \eqref{eq5} with the Sinkhorn distance\cite{cuturi2013sinkhorn}, we can get a new optimization objective
\begin{equation}
    \label{eq16}
    \begin{aligned}
        \mathop{\text{minimize}}_{\theta,\phi}& ~~\langle \gamma, \mathcal{M}(\mathbf{Z}_{\theta} ,\mathbf{Z}_T) \rangle_F + \epsilon\sum_{i,j}\gamma_{ij}\log(\gamma_{ij}) \\
        &~~+ \frac{\lambda}{n}\sum_{i=1}^n \|\mathbf{x}_i-g_\phi(f_\theta(\mathbf{x}_i))\|^2\\ 
        \text{subject to} & ~~ \gamma \mathbf{1} = \mathbf{a}, \gamma^T \mathbf{1} = \mathbf{b}, \gamma  \geq 0
    \end{aligned}
\end{equation}
where $\mathcal{M}(\mathbf{Z}_{\theta} ,\mathbf{Z}_T)$ denotes the metric cost matrix between $\mathbf{Z}_{\theta}$ and $\mathbf{Z}_T$, $\epsilon$ is the coefficient of entropic regularization term, $\mathbf{a}$ and $\mathbf{b}$ are two probability vectors and satisfy $\mathbf{a}^T\mathbf{1}=1$ and $ \mathbf{b}^T\mathbf{1}=1$ respectively. We call this method \textbf{Sinkhorn RGP}.

Compared to MMD, Sinkhorn distance is more effective in quantifying the difference between two distributions using their finite samples. Therefore, the Sinkhorn RGP usually has better performance than our original RGP \eqref{eq5}, which will be shown by the experimental results in Section~\ref{exp4.5}.

\section{Experiments}
\label{experiments}

\subsection{Datasets and Baselines}\label{sec_data_baseline}

We compare the proposed method with several state-of-the-art methods of anomaly detection on five tabular datasets and three widely-used image datasets for one-class classification. The datasets are detailed as follows.

\begin{itemize}
\item{\textbf{Abalone}\footnote{http://archive.ics.uci.edu/ml/datasets/Abalone}\cite{dua2017} is a dataset of physical measurements of abalone to predict the age. It contains 1,920 instances with 8 attributes.}
\item{\textbf{Arrhythmia}\footnote{http://odds.cs.stonybrook.edu/arrhythmia-dataset/}\cite{rayana2016} is an ECG dataset. It was used to identify arrhythmic samples in five classes and contains 452 instances with 279 attributes.} \item{\textbf{Thyroid}\footnote{http://odds.cs.stonybrook.edu/thyroid-disease-dataset/}\cite{rayana2016} is a hypothyroid disease dataset that contains 3,772 instances with 6 attributes.} \item{\textbf{KDD}\footnote{https://kdd.ics.uci.edu/databases/kddcup99/}\cite{Lichman2013KDD} is the KDDCUP99 10 percent dataset from the UCI repository and contains 34 continuous attributes and 7 categorical attributes. The attack samples are regarded as normal data, and the non-attack samples are regarded as abnormal data.}
\item{\textbf{KDDRev} is derived from the KDDCUP99 10 percent dataset. The non-attack samples are regarded as normal data, and the attack samples are regarded as abnormal data.} 
\item{\textbf{MNIST}\footnote{http://yann.lecun.com/exdb/mnist/}\cite{LeCun2010mnist} is a well-known dataset of handwritten digits and totally contains 70,000 grey-scale images in 10 classes from number 0-9.}
\item{\textbf{Fashion-MNIST}\footnote{https://www.kaggle.com/datasets/zalando-research/fashionmnist}\cite{xiao2017/online} contains 70,000 grey-scale fashion images (e.g. T-shirt and bag) in 10 classes.}
\item{\textbf{CIFAR-10}\footnote{https://www.cs.toronto.edu/ kriz/cifar.html}\cite{krizhevsky2009learning} is a widely-used benchmark for image anomaly detection. It contains 60,000 color images in 10 classes.}
\end{itemize}

We compare our method with three classic shallow models, four deep autoencoder based methods, three deep generative model based methods, and some latest anomaly detection methods.

\begin{itemize}
    \item{\textbf{Classic shallow models}: local outlier factor (LOF)\cite{breunig2000lof}, one-class support vector machine (OC-SVM)\cite{scholkopf2001estimating}, isolation forest (IF)\cite{liu2008isolation}.}
    
    \item{\textbf{Deep autoencoder based methods}: denoising auto-encoder (DAE)\cite{vincent2008extracting}, DCAE\cite{seebock2016identifying}, E2E-AE, DAGMM\cite{zong2018deep}, DCN \cite{caron2018deep}.}
    
    \item{\textbf{Deep generative model based methods}: AnoGAN\cite{schlegl2017unsupervised}, ADGAN\cite{deecke2018image}, OCGAN \cite{perera2019ocgan}.}
    
    \item{\textbf{Some latest AD methods}: DeepSVDD\cite{ruff2018deep}, GOAD \cite{bergman2019classification}, DROCC \cite{goyal2020drocc}, HRN \cite{hu2020hrn}, SCADN \cite{yan2021learning}, NeuTraL AD \cite{qiu2021neural}, GOCC \cite{shenkar2022anomaly}, PLAD \cite{cai2022plad}, MOCCA \cite{massoli2022mocca}.}
\end{itemize}

\begin{table*}[h!]
     \caption{Average AUC(\%) of one-class anomaly detection on Fashion-MNIST. For the competitive methods we only report their mean performance due to the space limit, while we further report the standard deviation for the proposed methods. `*' denotes we run the official released code to obtain the results, and the best two results are marked in \textbf{bold}.}
    \label{table1}
    \centering
    \begin{tabular}{l|cccccccccc}
    \toprule
        Normal Class & T-shirt & Trouser & Pullover & Dress & Coat & Sandal & Shirt & Sneaker & Bag & \makecell[c]{Ankle- \\ boot} \\
    \midrule
        OC-SVM\cite{scholkopf2001estimating} & 86.10 & 93.90 & 85.60 & 85.90 & 84.60 & 81.30 & 78.60 & 97.60 & 79.50 & 97.80\\
        IF\cite{liu2008isolation} & 91.00& 97.80 &87.20 & 93.20 &90.50 & 93.00 & 80.20 & 98.20 & 88.70 & 95.40\\
        DAE\cite{vincent2008extracting} & 86.70& 97.80 & 80.80 & 91.40 & 86.50 & 92.10 & 73.80 & 97.70  & 78.20 & 96.30\\
        DAGMM\cite{zong2018deep} & 42.10 & 55.10 & 50.40 & 57.00 & 26.90 & 70.50 & 48.30 & 83.50 & 49.90 & 34.00\\
        ADGAN\cite{deecke2018image} & 89.90 & 81.90 & 87.60 & 91.20 & 86.50 & 89.60 & 74.30 & 97.20 & 89.00 & 97.10\\
        OCGAN\cite{perera2019ocgan} &85.50 & 93.40& 85.00& 88.10 & 85.80 & 88.50 & 77.50 & 93.90 & 82.70 & 97.80\\
        DeepSVDD\cite{ruff2018deep} & 79.10 & 94.00& 83.00 & 82.90 & 87.00& 80.30& 74.90& 94.20 & 79.10 & 93.20\\
        DROCC$^*$\cite{goyal2020drocc} & 88.32 & 97.94 & 87.31 & 87.89 & 86.53 & 91.80 & 77.64 & 95.37 & 81.35 & 94.75\\
        HRN\cite{hu2020hrn} &92.70 & 98.50 & 88.50& 93.10 & 92.10 & 91.30 & 79.80 & \textbf{99.00} & \textbf{94.60} & \textbf{98.80}\\
        PLAD\cite{cai2022plad} & \textbf{93.10} & \textbf{98.60} & 90.20 & 93.70 & \textbf{92.80} & 96.00 & 82.00 & 98.60 & 90.90 & \textbf{99.10}\\
        \midrule    
        RGP-{GiHS} (Ours) & \makecell[c]{92.79 \\ (0.40)} & \makecell[c]{98.10 \\ (0.27)} & \makecell[c]{\textbf{90.45} \\(1.28)} & \makecell[c]{94.30 \\ (0.57)} & \makecell[c]{91.71 \\(0.30)} & \makecell[c]{\textbf{96.09} \\(0.67)} & \makecell[c]{85.91 \\ (0.39)} & \makecell[c]{98.58\\(0.08)} & \makecell[c]{92.67 \\ (1.10)} & \makecell[c]{97.11 \\ (0.23)}\\
        RGP-{UiHS} (Ours) & \makecell[c]{92.48 \\ (0.78)} & \makecell[c]{98.31\\(0.19)} & \makecell[c]{89.81\\(1.19)} & \makecell[c]{94.81\\(0.74)} & \makecell[c]{89.30\\(1.95)} & \makecell[c]{95.75\\(0.24)} & \makecell[c]{\textbf{85.95}\\(0.59)} & \makecell[c]{98.54\\(0.08)} & \makecell[c]{92.25\\(0.79)} & \makecell[c]{94.00\\(1.10)}\\
        RGP-{UbHS} (Ours) & \makecell[c]{92.83\\(0.68)} & \makecell[c]{97.88\\(0.61)} & \makecell[c]{90.19\\(1.02)} & \makecell[c]{\textbf{94.87}\\(0.34)} & \makecell[c]{91.97\\(0.78)} & \makecell[c]{\textbf{96.32}\\(0.18)} & \makecell[c]{85.76\\(0.48)} & \makecell[c]{98.67\\(0.13)} & \makecell[c]{91.32\\(1.05)} & \makecell[c]{94.93\\(1.00)}\\
        RGP-{UoHS} (Ours) & \makecell[c]{\textbf{94.85}\\(0.18)} & \makecell[c]{\textbf{98.94}\\(0.09)} & \makecell[c]{\textbf{92.39}\\(0.24)} & \makecell[c]{\textbf{95.71}\\(0.33)} & \makecell[c]{\textbf{93.12}\\(0.39)} & \makecell[c]{94.71\\(0.65)} & \makecell[c]{\textbf{86.98}\\(0.33)} & \makecell[c]{\textbf{99.16}\\(0.11)} & \makecell[c]{\textbf{94.16}\\(0.25)} & \makecell[c]{97.45\\(0.71)} \\
    \bottomrule

    \end{tabular}
    
\end{table*}
\begin{table*}[h!]
     \caption{Average AUC(\%) of one-class anomaly detection on CIFAR-10. More detailed description are the same as that in Table \ref{table1}.}
        \label{table2}
        \centering
    \begin{tabular}{l|cccccccccc}
       \toprule
        Normal Class & Airplane & \makecell[c]{Auto- \\ mobile} & Bird & Cat & Deer & Dog & Frog & Horse & Ship & Trunk \\
        \midrule
        OC-SVM\cite{scholkopf2001estimating} & 61.10 & 63.80  & 50.00&  55.90& 66.00&62.40 &74.70 & 62.60 & 74.90 & 75.90  \\
        IF\cite{liu2008isolation} & 66.10& 43.70 & 64.30 & 50.50 & \textbf{74.30} & 52.30 & 70.70 & 53.00 & 69.10 & 53.20  \\
        DCAE\cite{seebock2016identifying} & 59.10 & 57.40 & 48.90 & 58.40 & 54.00 & 62.20 & 51.20 & 58.60 & 76.80 & 67.30  \\
        DAE\cite{vincent2008extracting} & 41.10& 47.80 & 61.60 & 56.20 & \textbf{72.80} & 51.30 & 68.80 & 49.70 & 48.70 & 37.80  \\
        DAGMM\cite{zong2018deep} &41.40 & 57.10 & 53.80 & 51.20  &52.20 & 49.30&64.90 &55.30 &  51.90 & 54.20\\
        AnoGAN\cite{schlegl2017unsupervised} & 67.10 &  54.70& 52.90& 54.50 & 65.10 & 60.30 & 58.50 & 62.50 & 75.80 & 66.50  \\
        ADGAN\cite{deecke2018image} & 63.20& 52.90 & 58.00 & 60.60 & 60.70 & 65.90 & 61.10 & 63.00 & 74.40 & 64.20  \\
        OCGAN\cite{perera2019ocgan} & 75.70& 53.10 &  64.00& 62.00& 72.30 & 62.00 & 72.30 & 57.50 & 82.00 & 55.40 \\
        DeepSVDD\cite{ruff2018deep} &61.70 & 65.90 & 50.80 & 59.10 & 60.90 & 65.70 & 67.70 & 67.30 & 75.90 & 73.10  \\

        $\text{DROCC}^*$\cite{goyal2020drocc} & \textbf{80.10} & \textbf{73.41} & \textbf{68.78} & 63.36 & 70.81 & 65.01 & 68.83 & \textbf{71.13} & 63.81 & 75.49  \\

        HRN\cite{hu2020hrn} & 77.30  & 69.90 &  60.60 & 64.40 & 71.50 & 67.40 & 77.40 &  64.90& \textbf{82.50} & \textbf{77.30}  \\
        $\text{MOCCA}_{(h)}$\cite{massoli2022mocca} & 66.00 & 70.50 & 52.40 & 60.10 & 60.90 & \textbf{68.40} & 67.10 & 68.50 & 79.20 & 75.80\\
        $\text{MOCCA}_{(s)}$\cite{massoli2022mocca} & 62.60 & \textbf{74.60} & 57.50 & 57.80 & 61.50 & 66.30 & 67.40 & \textbf{72.10} & 79.10 & \textbf{77.30} \\
        \midrule
        RGP-{GiHS} (Ours) & \makecell[c]{77.01 \\ (0.61)}&\makecell[c]{68.56\\(0.34)} & \makecell[c]{62.57\\(0.82)}& \makecell[c]{63.06 \\ (0.29)} & \makecell[c]{70.72\\(1.28)} & \makecell[c]{\textbf{68.78}\\(0.76)}& \makecell[c]{\textbf{80.51}\\(0.95)}& \makecell[c]{67.92\\(0.61)}& \makecell[c]{80.50\\(1.13)}& \makecell[c]{73.06\\(1.30)}  \\
        
        RGP-{UiHS} (Ours) &\makecell[c]{76.07\\(1.92)} &\makecell[c]{70.66 \\ (0.23)}& \makecell[c]{\textbf{67.20}\\(0.34)}& \makecell[c]{\textbf{64.72}\\(2.67)} & \makecell[c]{70.38\\(0.51)} & \makecell[c]{67.63\\(0.39)}& \makecell[c]{80.25\\(0.94)}& \makecell[c]{69.44\\(0.82)}& \makecell[c]{81.19\\(0.96)}& \makecell[c]{74.89\\(0.24)}  \\
        
        RGP-{UbHS} (Ours) & \makecell[c]{77.66\\(0.37)}&\makecell[c]{68.76 \\ (1.23)} & \makecell[c]{65.29\\(0.32)}& \makecell[c]{64.40\\(1.56)} & \makecell[c]{69.89\\(1.16)} & \makecell[c]{68.00\\(0.95)}& \makecell[c]{\textbf{80.75} \\(0.18)}& \makecell[c]{68.79\\(0.75)}& \makecell[c]{82.17\\(0.56)}& \makecell[c]{73.87\\(0.81)}  \\
        
        RGP-{UoHS} (Ours) & \makecell[c]{\textbf{78.09} \\ (0.98)}&\makecell[c]{67.71\\(0.64)} & \makecell[c]{61.07\\(0.95)}& \makecell[c]{\textbf{66.48} \\ (0.30)} & \makecell[c]{69.70 \\ (0.22)} & \makecell[c]{68.37\\(0.66)}& \makecell[c]{80.14\\(0.66)}&\makecell[c]{70.9\\(0.37)}& \makecell[c]{\textbf{83.27}\\(0.28)}& \makecell[c]{74.10\\(0.46)}  \\
        \bottomrule
    \end{tabular}
    
\end{table*}

\subsection{Implementation Details and Evaluation Metrics} 
In this section, we introduce the implementation details of the proposed method RGP and describe experimental settings for image and tabular datasets. Note that our method neither uses any abnormal data during the training process nor utilizes any pre-trained feature extractors.

For the five tabular datasets (Abalone, Arrhythmia, Thyroid, KDD, KDDRev), in our method, $f_\theta$ and $ g_\phi$ are both MLPs. We follow the dataset preparation of ~\cite{zong2018deep} to preprocess the tabular datasets for one-class classification task. The hyper-parameter $\lambda$ is set to 1.0 for the Abalone, Arrhythmia and Thyroid. For the KDD and KDDRev, $\lambda$ is set to 0.0001. 

For the three image datasets (MNIST, Fashion-MNIST, CIFAR-10), in our method, $f_\theta$ and $g_\phi$ are both CNNs. Since the three image datasets contain 10 different classes, we conduct 10 independent one-class classification tasks on both datasets: one class is regarded as normal data and the remaining nine classes are regarded as abnormal data. In each task on MNIST, there are about 6,000 training samples and 10000 testing samples. In each task on CIFAR-10, there are 5,000 training samples and 10,000 testing samples. In each task on Fashion-MNIST, there are 6,000 training samples and 10,000 testing samples. The hyper-parameter $\lambda$ is chosen from $\{1.0, 0.5, 0.1, 0.01, 0.001, 0.0001\}$ and varies for different classes.

In our method, regarding the radius $r$ of GiHS and UiHS, we first generate a large number (denoted by $N$) of samples from Gaussian or uniform, sort the samples according to their $\ell_2$ norms, and set $r$ to be the $pN$-th smallest $\ell_2$ norm, where $p=0.9$. For UbHS, we need to use the aforementioned method to determine an $r$ with $p=0.95$ and a $r'$ with $p=0.05$. We see that $\{r, r'\}$ are not related to the actual data, they are determined purely by the target distribution.
In each iteration (mini-batch) of the optimization for all four target distributions, we resample $\mathbf{Z}_T$ according to $r$. For UoHS, we draw samples from Gaussian and normalize them to have unit $\ell_2$ norm, then they lie on a unit hypersphere uniformly. The procedure is repeated in each iteration (mini-batch) of the optimization.
For hyper-parameter $k$ on the testing stage, we select $k=3$ for Thyroid, Arrhythmia, KDD, KDDRev, and select $k=5$ for Abalone dataset. For three image datasets, the hyper-parameter $k$ is chosen from $\{1, 3, 5, 10\}$ and varies for different classes.
We use Adam~\cite{kingma2014adam} as the optimizer in our method. For MNIST, Fashion-MNIST, CIFAR-10, Arrhythmia and KDD, the learning rate is set to $0.0001$. For Abalone, Thyroid and KDDRev, the learning rate is set to $0.001$. Table \ref{table17} shows the detailed implementation settings of RGP on all datasets. All experiments were run on AMD EPYC CPU with 64 cores and with NVIDIA Tesla A100 GPU, CUDA 11.6.

\begin{table}[h]
\caption{The detailed implementation settings of RGP on all datasets.}
\label{table17}
\begin{center}
\begin{tabular}{|l|c|c|c|}
\hline
Datasets & features & latent dimension & learning rate \\
\hline
Thyroid         & 6  & 4 & 0.001   \\
\hline
Abalone         & 8 & 4 & 0.001 \\
\hline
KDD             & 121 & 64 & 0.0001 \\
\hline
KDDRev          & 121 & 64 & 0.001 \\
\hline
Arrhythmia      & 279 & 128 & 0.0001 \\
\hline
MNIST           & 28$\times$28$\times$1 & 128 & 0.0001\\
\hline
Fashion-MNIST   & 28$\times$28$\times$1 & 128 & 0.0001\\
\hline
CIFAR-10        & 32$\times$32$\times$3 & 128 & 0.0001\\
\hline
\end{tabular}

\end{center}
\end{table}

To evaluate the performance of all methods, we follow the previous works such as ~\cite{ruff2018deep} and ~\cite{zong2018deep} to use AUC (Area Under the ROC curve) for image datasets and F1-score for tabular datasets.
Note that when conducting experiments on the tabular datasets, we found that most of the strong baselines, like DROCC~\cite{goyal2020drocc}, NeuTral AD~\cite{qiu2021neural}, GOCC~\cite{shenkar2022anomaly}, used the F1-score and we just followed this convention.
In our method, we get the threshold via simply calculating the dispersion of training data in latent space. Specifically, we first calculated the scores $s(\mathbf{X})$ on training data $\mathbf{X}$ using (12) or (13), and then sorted $s(\mathbf{X})$ in ascending order and set the threshold to be the $pN$-th smallest score, where $p$ is a probability varying for different datasets.

\subsection{Results on Image Datasets}
Tables~\ref{table1} and~\ref{table2} show the comparison results on Fahsion-MNIST and CIFAR-10 respectively. We have the following observations.

\begin{itemize}
    \item Firstly, in contrast to classic shallow methods such as OC-SVM \cite{scholkopf2001estimating} and IF~\cite{liu2008isolation},  our RGP has significantly higher AUC scores on all classes of Fashion-MNIST and most classes of CIFAR-10. An interesting phenomenon is that most deep learning based methods have inferior performance compared to IF~\cite{liu2008isolation} on class `Sandal' of Fashion-MNIST and IF~\cite{liu2008isolation} outperforms all deep learning based methods including ours on class `Deer' of CIFAR-10. 
    \item Our methods outperformed the deep autoencoder based methods and generative model based methods in most cases and have competitive performance compared to the state-of-the-art in all cases.
    \item RGP has superior performance on most classes of Fashion-MNIST and CIFAR-10 under the setting of UoHS (uniform distribution on hypersphere).
\end{itemize}

\begin{table}[h!]
\caption{Average AUC (\%) over all 10 classes of each image dataset. The best two results in each case are marked in \textbf{bold}.}
\label{table4}
\begin{center}
\begin{small}
\begin{tabular}{l|cccr}
\toprule
Methods & MNIST &  Fashion-MNIST & CIFAR-10 \\
\midrule
OC-SVM    & 91.28 & 87.09 & 64.72 \\
IF        & 92.29 & 91.52 & 59.72 \\
DAE & - & 88.13  & 53.57 \\
DAGMM & - & 51.77  & 53.13  \\
AnoGAN    & 91.27 & - & 61.79 \\
Deep SVDD & 94.79 & 84.77 & 64.81 \\
DROCC      & -  & 88.89& 70.07        \\
HRN   & \textbf{97.59} & 92.84 & 71.32 \\
$\text{MOCCA}_{(h)}$ & - & - & 66.90\\
$\text{MOCCA}_{(s)}$ & - & - & 67.60\\
\midrule
RGP-{GiHS} & 93.75 & \textbf{93.77} & 71.26\\
RGP-{UiHS} & 94.02 & 93.12 & \textbf{72.24} \\
RGP-{UbHS} & 93.60 & 93.47 & 71.97\\
RGP-{UoHS} & \textbf{95.81} & \textbf{94.74} & \textbf{71.98}\\
\bottomrule
\end{tabular}
\end{small}
\end{center}
\end{table}

Table \ref{table4} shows the average performance on MNIST, Fashion-MNIST, and CIFAR-10 over all 10 classes to provide an overall comparison. We see that RGP achieves the best average AUC on Fashion-MNSIT and CIFAR-10 among all competitive methods. Four variants of RGP have relatively close average performance on all three image datasets. The experimental results of a single class on MNIST are reported in Appendix.

\begin{table*}[h!]
    \centering
     \caption{Average F1-Scores(\%) with standard deviation on five tabular datasets. `*' denotes we run the officially released code of NeuTral AD to obtain the result of Abalone, and the results of Arrhythmia and Thyroid are from the original paper ~\cite{qiu2021neural}. The best two results are marked in \textbf{bold}.}
     \label{table3}
    \begin{tabular}{l|ccccc}
       \toprule
        Methods & Abalone & Arrhythmia & Thyroid & KDD & KDDRev\\
        \midrule
        OC-SVM~\cite{scholkopf2001estimating} & 48.00 $\pm$ 0.00 & 46.00 $\pm$ 0.00 & 39.00 $\pm$ 1.00 & 79.50 & 83.20\\
        LOF~\cite{breunig2000lof} &  33.00 $\pm$ 1.00 & 51.00 $\pm$ 1.00 &  54.00 $\pm$ 1.00 & 83.80 & 90.60 \\
        DCN~\cite{caron2018deep} & 40.00 $\pm$ 1.00 & 38.00 $\pm$ 3.00 & 33.00 $\pm$ 3.00 & - & -\\
        E2E-AE~\cite{zong2018deep} & 33.00 $\pm$ 3.00 & 45.00 $\pm$ 3.00 & 13.00 $\pm$ 4.00 & - & -\\ 
        DAGMM~\cite{zong2018deep} & 20.00 $\pm$ 3.00 & 49.00 $\pm$ 3.00 & 49.00 $\pm$ 4.00  & 93.70 & 93.80\\
        DeepSVDD~\cite{ruff2018deep} & 62.00 $\pm$ 1.00 & 54.00 $\pm$ 1.00 & 73.00 $\pm$ 0.00 & 99.00 $\pm$ 0.10 & 98.60 $\pm$ 0.20 \\
        GoAD~\cite{bergman2019classification} & 61.00 $\pm$ 2.00 & 51.00 $\pm$ 2.00  & 72.00 $\pm$ 1.00 & 98.40 $\pm$ 0.20 & 98.90 $\pm$ 0.30\\
        DROCC~\cite{goyal2020drocc} & 68.00 $\pm$ 2.00 & 69.00 $\pm$ 2.00 & 78.00 $\pm$ 3.00 & - & -\\
        NeuTral AD$^*$~\cite{qiu2021neural} & 62.07 $\pm$ 2.81  & 60.30 $\pm$ 1.10 & 76.80 $\pm$ 1.90 & 99.30 $\pm$ 0.10 & \textbf{99.10} $\pm$ 0.10\\
        GOCC ~\cite{shenkar2022anomaly} & -  & 61.80 $\pm$ 1.80 & 76.80 $\pm$ 1.20 & \textbf{99.40} $\pm$ 0.10 & \textbf{99.20} $\pm$ 0.30 \\
        \midrule
        RGP-{GiHS} (Ours) & \textbf{91.25} $\pm$ 1.92 & \textbf{81.22} $\pm$ 0.50 & \textbf{97.58} $\pm$ 0.48 & 99.29 $\pm$ 0.10 & 98.99 $\pm$ 0.02\\
        RGP-{UiHS} (Ours) & \textbf{90.38} $\pm$ 1.87 & \textbf{81.02} $\pm$ 0.81 &  97.09 $\pm$ 0.27 & 99.28 $\pm$ 0.19 & 98.96 $\pm$ 0.07\\
        RGP-{UbHS} (Ours) & 90.20 $\pm$ 2.32 & 81.00 $\pm$ 0.67 &  97.17 $\pm$ 0.55 & 99.13 $\pm$ 0.31 & 98.99 $\pm$ 0.03\\
        RGP-{UoHS} (Ours) & 89.59 $\pm$ 1.52 & 80.97 $\pm$ 0.62 &  \textbf{97.38} $\pm$ 0.36 & \textbf{99.43} $\pm$ 0.01 & 99.07 $\pm$ 0.03\\
        \bottomrule
    \end{tabular}
\end{table*}

\subsection{Results on Tabular Datasets}
In Table \ref{table3}, we report the F1-scores of our methods in comparison to ten baselines on the five tabular datasets. Our four variants of RGP significantly outperform all baseline methods on Arrhythmia, thyroid, and Abalone. Particularly, RGP-GiHS has $23.25\%$, $12.22\%$, and $19.58\%$ improvements on the three datasets in terms of F1-score compared to the runner-up, respectively. It is worth mentioning that Neutral AD \cite{qiu2021neural} and GOCC \cite{shenkar2022anomaly} are both specially designed for non-image data but are outperformed by our methods in most cases.
Compared with image datasets, the performance improvements of RGPs on the three tabular datasets are more significant. One possible reason is that, compared to image data, it is easier to convert tabular data to a compact target distribution. Furthermore, we also report the AUC scores on Abalone, Thyroid and Arrhythmia datasets and the results are provided in Appendix.

\begin{figure}[h!]
\centering
    \includegraphics[width=0.45\textwidth]{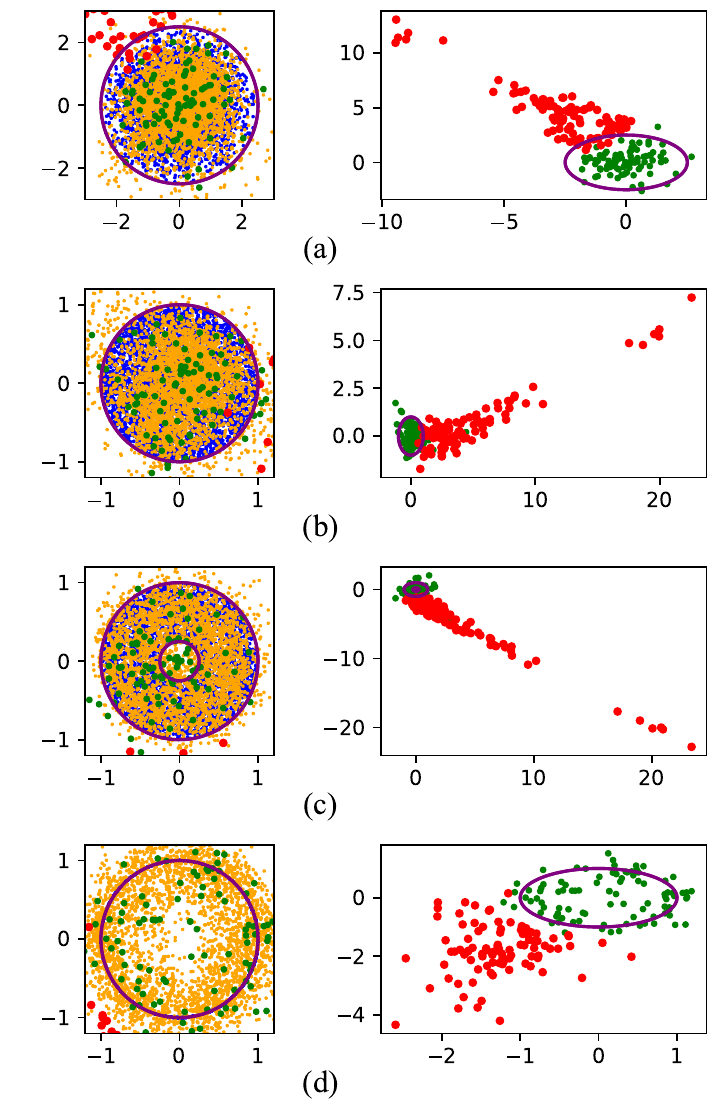}
 \caption{Visualization of mapping data distribution to 2-dimensional target distribution on Thyroid datasets. Plots (a), (b), (c), (d) refer to \textbf{GiHS}, \textbf{UiHS}, \textbf{UbHS}, \textbf{UoHS}, respectively. For a clear illustration, the left figure in each plot shows all four kinds of instances and the right figure shows two kinds of instances including normal and abnormal samples from test set.}\label{fig3}
\end{figure}

In addition to the quantitative results, we choose Thyroid (with 6 attributes) as an example and transform the data distribution to 2-dimensional target distributions, which are visualized in Figure~\ref{fig3}. Plots (a), (b), (c), (d) in Figure~\ref{fig3} refer to \textbf{GiHS}, \textbf{UiHS}, \textbf{UbHS}, \textbf{UoHS}, respectively. The blue points, orange points, green points, and red points denote samples from target distribution, samples from training data, normal samples from test set, and abnormal samples from test set, respectively. For much clearer illustration, the left figure in each plot of Figure~\ref{fig3} shows all four kinds of instances and the right figure shows two kinds of instances including normal and abnormal samples from test set.
We see that RGPs are effective to transform the data distribution to the restricted target distributions, though the transformed data do not exactly match the target distributions (it also demonstrates the necessity of using the `soft boundary' defined by \eqref{eq13}).

\subsection{Comparison between`soft' and `hard' boundary}
\label{section4.6}
\begin{figure}[h!]
    \centering
    \includegraphics[width=0.5\textwidth]{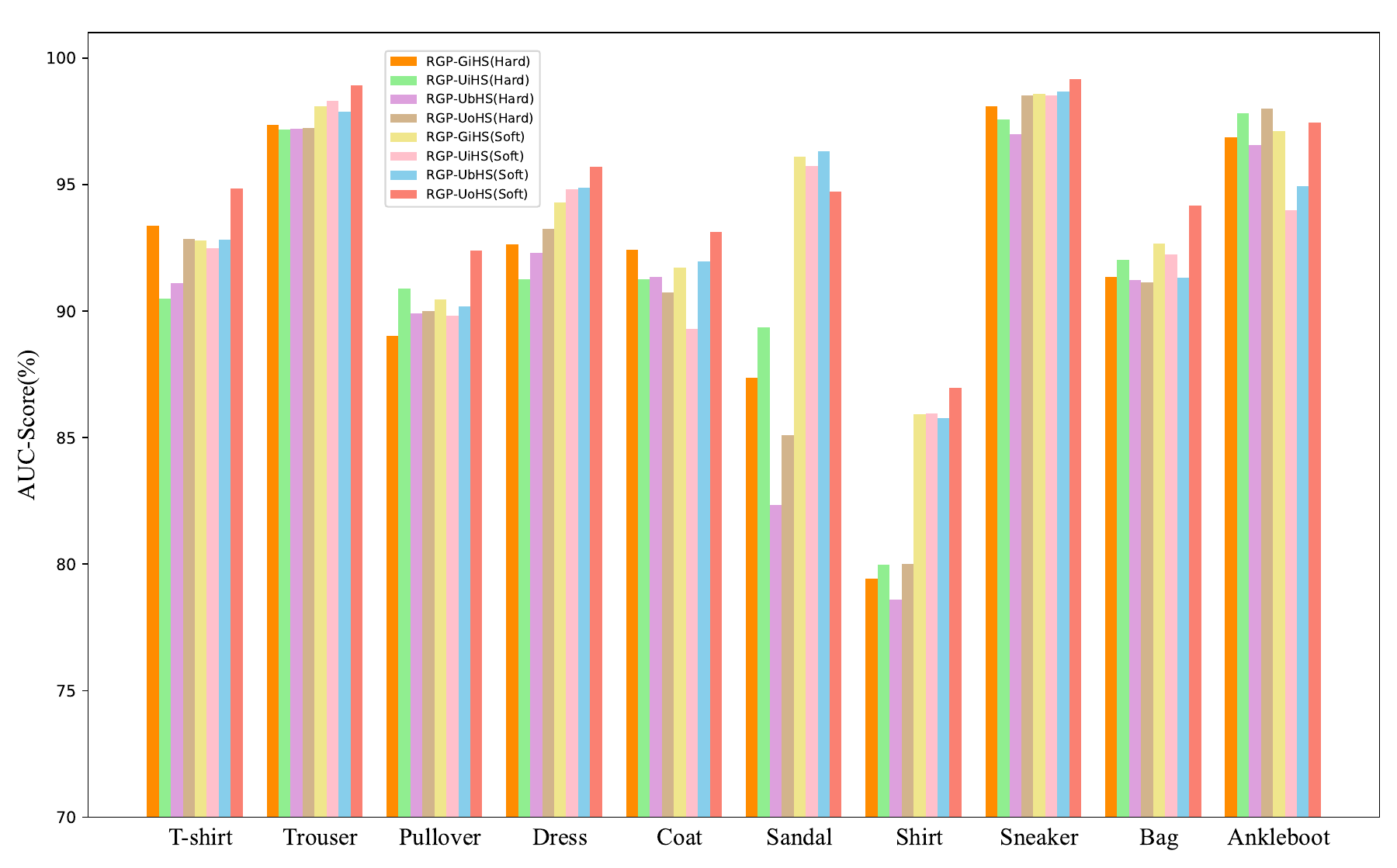}
    \caption{Performance comparison between `hard boundaries' and `soft boundary' as anomaly score of RGPs on Fashion-MNSIT.}
    \label{fig4_1}
\end{figure}

\begin{figure}[h!]
    \centering
    \includegraphics[width=0.5\textwidth]{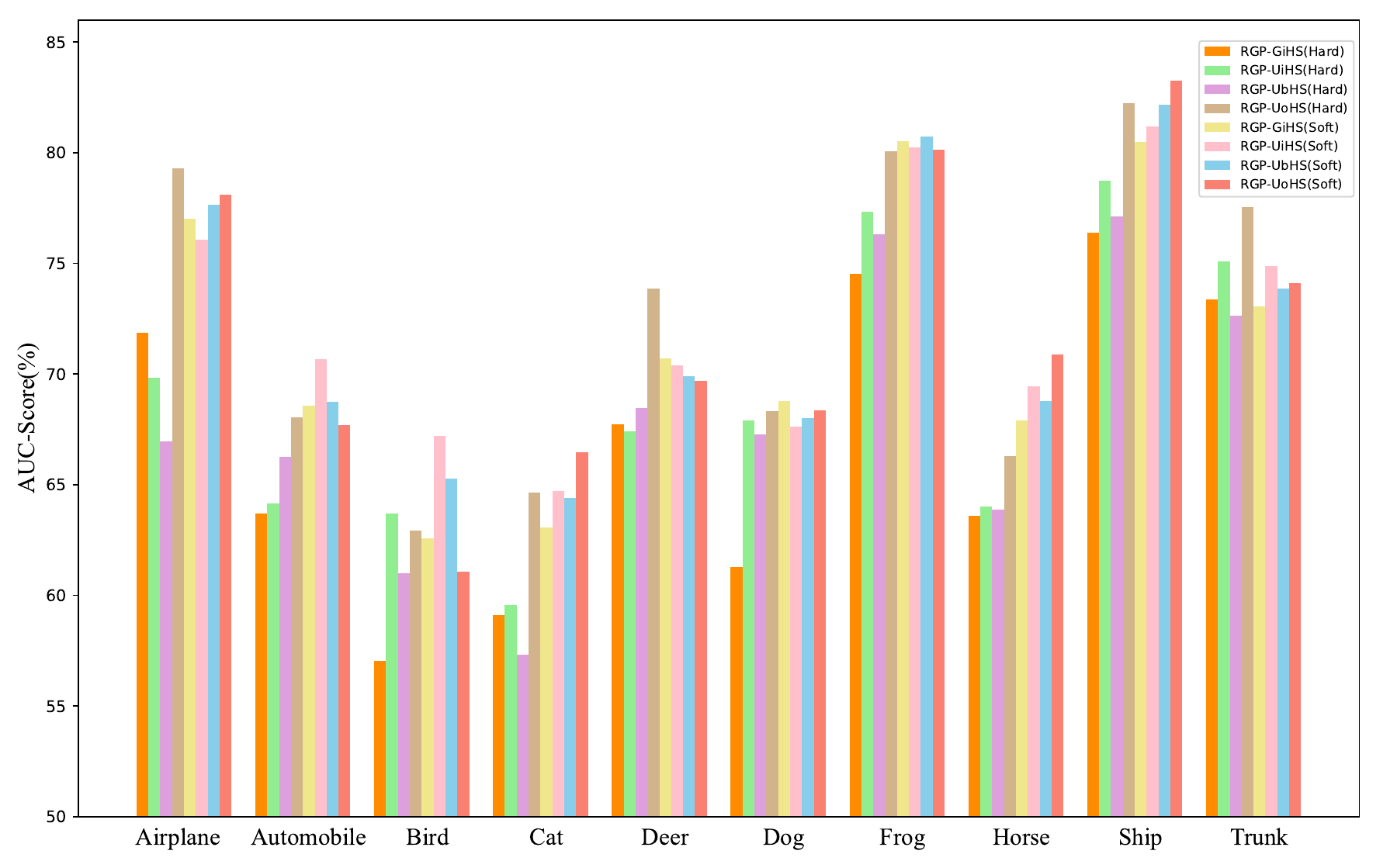}
    \caption{Performance comparison between `hard boundaries' and `soft boundary' as anomaly score of RGPs on CIFAR-10.}
    \label{fig5_1}
\end{figure}

\begin{figure}[h!]
    \centering
    \includegraphics[width=0.5\textwidth]{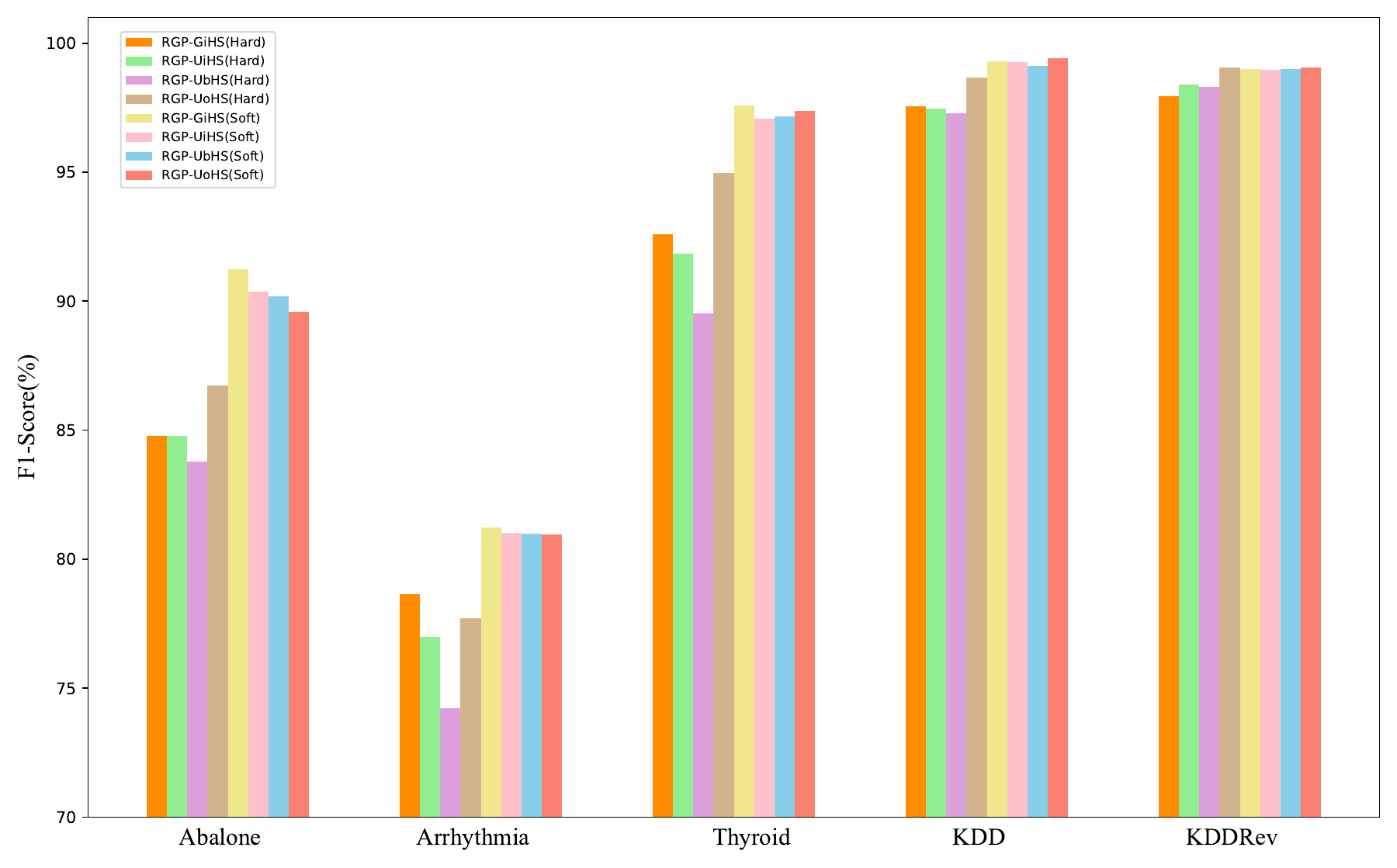}
    \caption{Performance comparison between `hard boundaries' and `soft boundary' as anomaly score of RGPs on five tabular datasets.}
    \label{fig6_1}
\end{figure}

We further explore the performance of two different anomaly scores. Specifically, we compare the `hard boundaries' (\ref{eq12}) and `soft boundary' (\ref{eq13}) as anomaly scores during the test stage on image datasets and tabular datasets. The results are showed in Figures~\ref{fig4_1}, \ref{fig5_1}, \ref{fig6_1}. It can be observed that using `soft boundary' (\ref{eq13}) to calculate anomaly score has better performance than using `hard boundaries' (\ref{eq12}) on most classes of image and tabular datasets. Nevertheless, using `hard boundaries' to calculate anomaly scores still achieves remarkable performance on some classes. For example, on the class `Ankle-boot' of Fashion-MNIST and the class `Trunk' of CIFAR-10, the best two results are both from RGPs using `hard boundaries' (\ref{eq12}) to calculate anomaly score.


\subsection{Experiments of Double-MMD RGP and Sinkhorn RGP}
\label{exp4.5}
We use Double-MMD RGP~\eqref{eq15} to conduct experiments and the results are reported in Table~\ref{table8}, \ref{table10}. On image datasets, we just consider the target distribution UoHS (Uniform on HyperSphere) for simplicity.
On tabular datasets, we conduct experiments on the proposed four different target distributions.

From the experimental results of Table \ref{table8}, \ref{table10}, we found that Double-MMD RGP and original RGP have similar performance on the three tabular datasets, whereas on image datasets including Fashion-MNIST and CIFAR-10, the performance has apparent gap in spite of a large range of adjustment of $\lambda \in \{10.0, 5.0, 1.0, 0.5, 0.1, 0.01\}$ for Double-MMD RGP \eqref{eq15}. Note that Table~\ref{table8} reports the average AUC(\%) on all classes of Fahion-MNIST and CIFAR-10, the results on single class are provided in Appendix. 

\begin{table}[h]
        \caption{The Average AUC(\%) on all class of Fashion-MNIST and CIFAR-10 using two different optimization objectives under UoHS.\label{table8}}
    \centering
    \begin{tabular}{l|l|cc}
    \toprule
        & $\lambda$ & Fashion-MNIST & CIFAR-10 \\
    \midrule
    & $\lambda$=10.0 & 80.34 & 65.45\\
    & $\lambda$=5.0  & 77.23 & 66.34\\
    \textbf{Double-MMD RGP}~\eqref{eq15} & $\lambda$=1.0 & 79.95 & 66.60\\
    & $\lambda$=0.5  & 79.68 & 66.10\\
    & $\lambda$=0.1  & 79.08 & 69.08\\
    & $\lambda$=0.01 & 77.47 & 67.19\\
        \midrule
    \textbf{Original RGP}~\eqref{eq5} & & 94.74 & 71.98   \\
    \bottomrule
    \end{tabular}

\end{table}

For the phenomenon, we consider that the tabular datasets in our implementation have fewer features (no more than 279) than the image datasets and second term of \eqref{eq15} is a much weaker constraint for preserving data information than that of \eqref{eq5}. As a consequence, Double-MMD RGP~\eqref{eq15} is able to preserve the enough key information on the tabular data but loses a lot of important information on the image data than original RGP~\eqref{eq5}. Meanwhile, we know that the generalization error of MMD for high-dimensional samples or distribution is often larger than that for low-dimensional samples or distribution. To ensure that MMD is able to accurately measure the distance between two high-dimensional distributions, the sample sizes should be sufficiently large.

We use Sinkhorn RGP~\eqref{eq16} to conduct experiments on Abalone, Arrhythmia, and Thyroid datasets and the results are reported in Table~\ref{table10}. In all implementations, $\epsilon$ is set to $0.01$ and the  $\textbf{a, b}$ are uniform. In keeping with our expectation, the performance of Sinkhorn RGP~\eqref{eq16} is similar to or better than the original RGP~\eqref{eq5} for all four objective distributions, whereas the time cost of Sinkhorn RGP~\eqref{eq16} is much higher. We do not experiment with Sinkhorn RGP for the image dataset since the time cost is too higher.

\begin{table}[h]
    \centering
    \caption{The average AUC (\%) of Double-MMD RGP, Sinkhorn RGP, and original RGP on the tabular dataset. The best result of each optimization objective is marked in \textbf{bold}.\label{table10}}
    \begin{tabular}{l|c|ccc}
    \toprule
        & Datasets & Abalone & Arrhythmia & Thyroid \\
    \midrule
         & RGP-GiHS & 93.65 & 82.79 & 98.95 \\
    \textbf{Original RGP} & RGP-UiHS & \textbf{95.64} & \textbf{82.90} & \textbf{99.06} \\
         & RGP-UbHS & 94.93 & 82.70  & 98.92 \\
         & RGP-UoHS & 94.95 & 82.89 & 98.93 \\
    \midrule
         & RGP-GiHS & 95.19 & 81.51  & 98.94  \\
    \textbf{Sinkhorn RGP} & RGP-UiHS & 94.72 & 82.37 & 98.85 \\
         & RGP-UbHS & \textbf{95.41} & \textbf{83.31}  &  98.97\\
         & RGP-UoHS & 95.17 & 83.20 &  \textbf{98.99}\\
    \midrule
         & RGP-GiHS & \textbf{94.91} & 82.26 & 98.53\\
    \textbf{Double-MMD RGP} & RGP-UiHS & 94.83 & 82.19 & 98.69\\
         & RGP-UbHS & 93.88 & \textbf{82.28} & 98.73\\
         & RGP-UoHS & 92.60 & 80.73 & \textbf{98.89}\\
    \bottomrule
    \end{tabular}
\end{table}

\subsection{Ablation Study}
\subsubsection{\textbf{The Gaussian Kernel Function for MMD}}
We use the Gaussian kernel $\exp(-\gamma \Vert  \mathbf{x} - \mathbf{y}\Vert^2)$ for MMD in optimization objective and set $\gamma = \frac{1}{d^2}$ in all experiments, where $d=\frac{1}{n(n-1)} \sum^n_{i=1} \sum^n_{j=1} \Vert \mathbf{x}_i\ - \mathbf{x}_j \Vert$ denotes the mean Euclidean distance among all training samples. 

\begin{table*}[h!]
\centering
     \caption{The Comparison among different $\gamma$ in Gaussian kernel for MMD. `Avg' denotes the average performance on all ten classes.}
    \label{table12}
    \begin{tabular}{l|l|ccccccccccc}
    \toprule
        & Normal Class & T-shirt & Trouser & Pullover & Dress & Coat & Sandal & Shirt & Sneaker & Bag & \makecell[c]{Ankle- \\ boot} & Avg\\
    \midrule
    & $\gamma = 0.1 $ & 90.24 & 96.68 & 88.33 & 93.20 & 90.42 & 97.09 & 86.06 & 97.32 & 88.44 & 93.83 & 92.16\\
    GiHS & $\gamma = 1 $ & 90.73 & 98.22 & 89.08 & 92.90 & 88.12 & 94.70 & 87.15 & 98.24 & 90.24 & 98.40 & 92.77\\
    & $\gamma = 10 $ & 89.43 & 99.01 & 85.96 & 93.54 & 87.92 & 94.90 & 83.30 & 97.71 & 91.84 & 92.79 & 91.64\\
    & $\gamma = 100 $ & 92.84 & 98.26 & 84.80 & 95.50 & 86.69 & 95.16 & 86.36 & 98.75 & 86.78 & 95.83 & 92.09\\
    \midrule
    & $\gamma = 0.1 $ & 88.14 & 98.25 & 88.55 & 93.86 & 91.79 & 94.93 & 87.40 & 97.46 & 86.14 & 91.46 & 91.79\\
    UiHS & $\gamma = 1 $ & 90.49 & 98.48 & 90.05 & 92.77 & 92.57 & 95.07 & 85.11 & 98.17 & 88.23 & 94.60 & 92.55\\
    & $\gamma = 10 $ & 88.62 & 98.50 & 88.77 & 94.08 & 86.29 & 93.97 & 87.27 & 98.36 & 94.70 & 90.53 & 92.10\\
    & $\gamma = 100 $ & 88.62 & 98.50 & 88.77 & 94.08 & 86.29 & 93.97 & 87.27 & 98.36 & 94.70 & 90.53 & 92.10\\
    \bottomrule
    \end{tabular}
\end{table*}

To show the influence of $\gamma$,  we fix $\gamma$ from $\{0.1, 1, 10, 100\}$ to run experiments on Fashion-MNIST.
As shown in Table~\ref{table12}, there are differences in every single case but the gaps in the average results are not significant. This demonstrated that our methods are not sensitive to $\gamma$.

\subsubsection{\textbf{The Coefficient $\lambda$ of Reconstruction Term in Optimization Objective}}
\begin{figure}[h!]
\centering
    \includegraphics[width=0.4\textwidth]{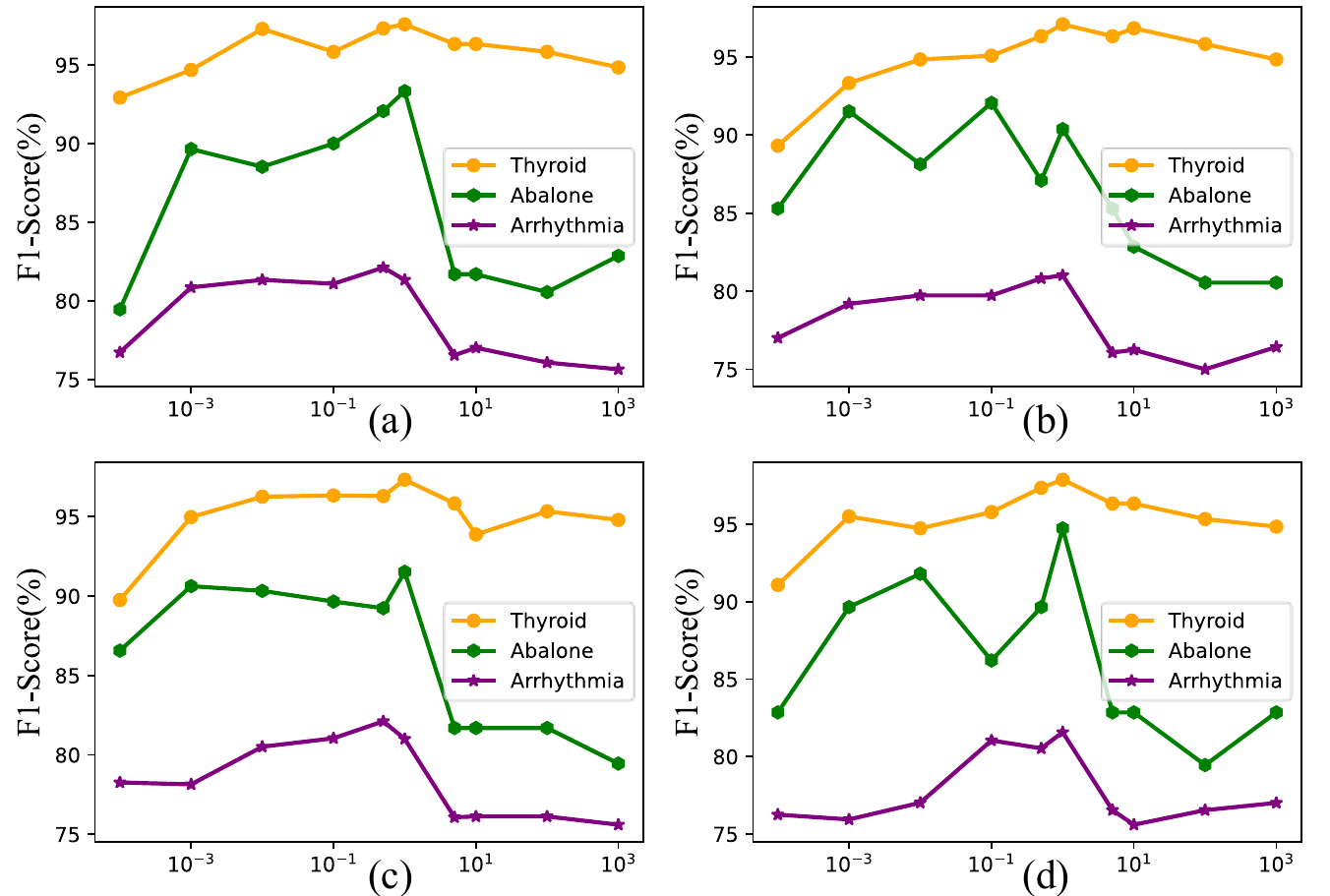}
\caption{The ablation study of hyper-parameter $\lambda$ on the test set of Abalone, Arrhythmia, Thyroid datasets under four different restrictions. Plots (a), (b), (c), (d) correspond to \textbf{GiHS}, \textbf{UiHS}, \textbf{UbHS}, \textbf{UoHB}, respectively. $\lambda$ is chosen from $\{0, 0.001, 0.01, 0.1, 0.5, 1, 5, 10, 100, 1000\}$.}\label{fig4}
\end{figure}

\begin{figure}[h!]
\centering
    \includegraphics[width=0.4\textwidth]{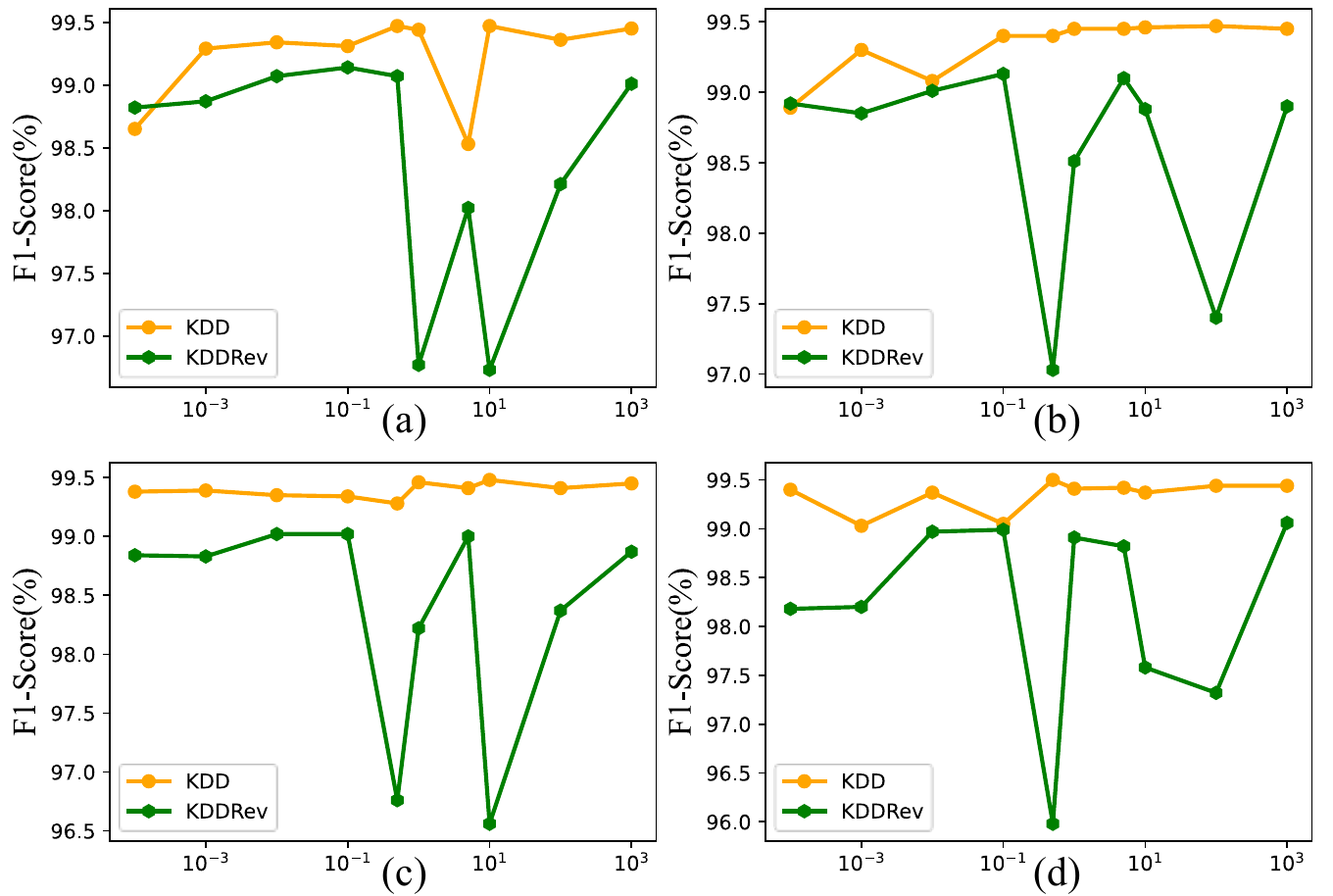}
\caption{The ablation study of hyper-parameter $\lambda$ on the test set of KDD and KDDRev datasets under four different restrictions. More detailed description are the same as that in Fig.\ref{fig4}.}\label{fig5}
\end{figure}

\begin{figure}[h!]
\centering
    \includegraphics[width=0.44\textwidth]{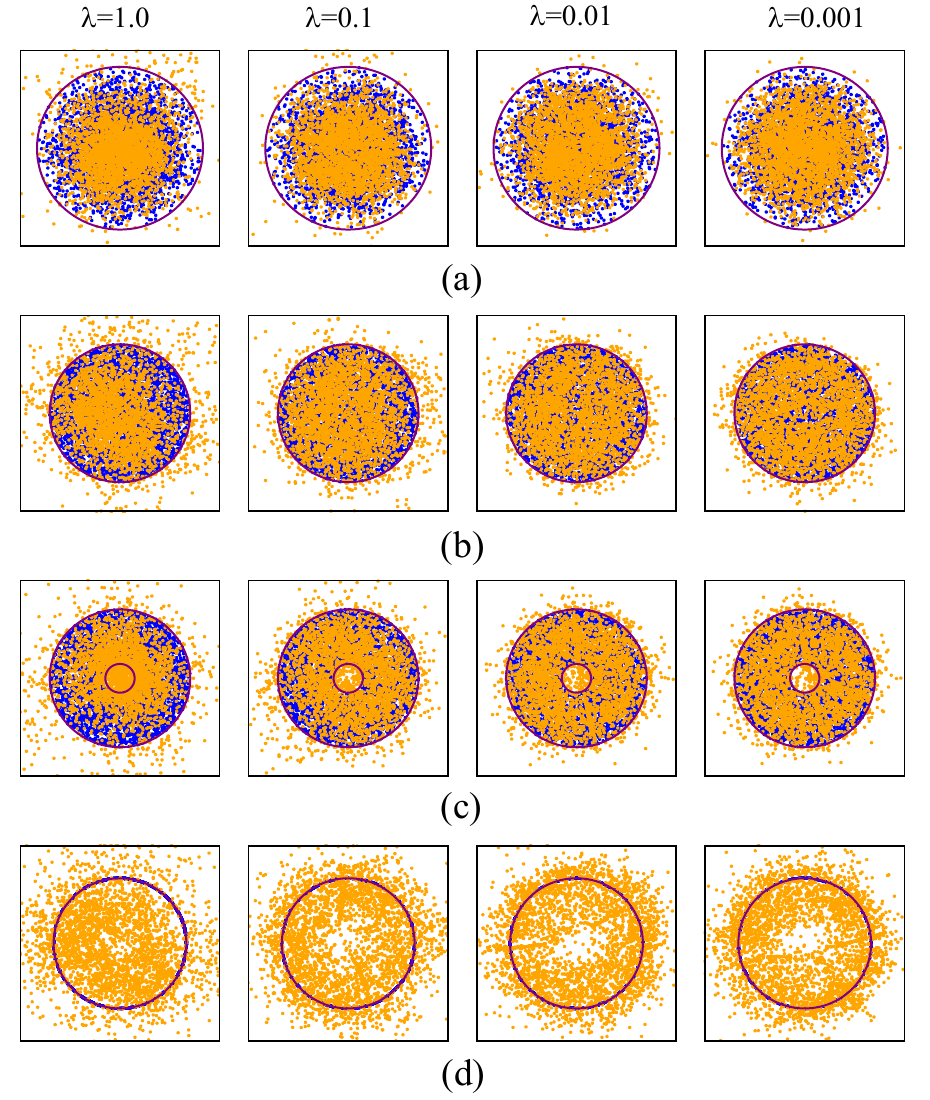}
\caption{The ablation study of hyper-parameter $\lambda$ on the training set of Thyroid dataset under four different restrictions using Sinkhorn RGP. Plots (a), (b), (c), (d) correspond to \textbf{GiHS}, \textbf{UiHS}, \textbf{UbHS}, \textbf{UoHS}, respectively. The blue and orange points denote samples from target distribution and samples from training data, respectively.}\label{fig6}
\end{figure}

The coefficient $\lambda$ is a key hyperparameter in problem (\ref{eq5}). Now we explore the influence of $\lambda$ for model performance. 
Figures~\ref{fig4},~\ref{fig5} show F1-scores of our methods with $\lambda$ varying from 0 to 1000, on the tabular datasets. It can be observed that too small or too large $\lambda$ can lower the performance of RGP. When $\lambda$ is very tiny, the reconstruction term of (\ref{eq5}) makes less impact on the training target and $f_\theta$ can easily transform the training data to the target distribution but ignores the importance of original data distribution (see Figure~\ref{fig6}). On the other hand, when $\lambda$ is very large, the MMD term of optimization objective becomes trivial for the whole training target and $f_\theta$ under the constraint of reconstruction term more concentrates on the original data distribution yet can not learn a good mapping from data distribution to the target distribution. Figure~\ref{fig6} illustrates the influence of hyper-parameter $\lambda$ on the training set of Thyroid dataset. We see that $f_\theta$ transforms training data to target distribution better with the decrease of the $\lambda$. The blue points and orange points in Figure~\ref{fig6} denote samples from target distribution, samples from training data, respectively.


\section{Conclusion}
\label{conclusion}
We have presented a novel and simple framework for one-class classification and anomaly detection. Our method aims to convert the data distribution to a simple, compact, and informative target distribution that can be easily violated by abnormal data. We presented four target distributions and the numerical results showed that four different target distributions have relatively close performance and uniform on hypersphere is more effective than other distributions in most cases. Furthermore, we also explore two extensions based on the original RGP and analyze performance difference among them. Importantly, our methods have competitive performances as state-of-the-art AD methods on all benchmark datasets considered in this paper and the improvements are remarkable on the tabular datasets.

\bibliographystyle{IEEEtran}
\bibliography{example_paper}

\vfill

\end{document}